\documentclass{article}

\PassOptionsToPackage{numbers, compress}{natbib}
\usepackage[final]{nips_2017}

\usepackage[utf8]{inputenc} 
\usepackage[T1]{fontenc}    
\usepackage{hyperref}       
\usepackage{url}            
\usepackage{booktabs}       
\usepackage{amsfonts}       
\usepackage{nicefrac}       
\usepackage{microtype}      

\usepackage{multirow}
\usepackage{listings}
\usepackage{graphicx}
\usepackage{comment}
\usepackage{subfigure}
\usepackage[dvipsnames]{xcolor}
\usepackage{color}
\usepackage{caption}

\newcommand{\HM}[1]{{\color{Orchid}{[}\textbf{Huizi: #1}{]}}}
\newcommand{\SH}[1]{{\color{Blue}{[}\textbf{Song: #1}{]}}}

\title{Exploring the Regularity of Sparse Structure in Convolutional Neural Networks}

\newcommand*{\affaddr}[1]{#1} 
\newcommand*{\affmark}[1][*]{\textsuperscript{#1}}
\newcommand*{\email}[1]{\texttt{#1}}
\author{
Huizi Mao\affmark[1], Song Han\affmark[1], Jeff Pool\affmark[2], Wenshuo Li\affmark[3], Xingyu Liu\affmark[1], \\
\textbf{Yu Wang\affmark[3], William J. Dally\affmark[1,2]}\\
\affaddr{\affmark[1]Stanford University}\\
\affaddr{\affmark[2]NVIDIA}\\
\affaddr{\affmark[3]Tsinghua University}\\
\email{\{huizi,songhan,dally\}@stanford.edu}
}

\begin{document}

\maketitle

\begin{abstract}
Sparsity helps reduce the computational complexity of deep neural networks by skipping zeros. Taking advantage of sparsity is listed as a high priority in the next generation DNN accelerators such as TPU\cite{Jouppi2017}. The structure of sparsity, i.e., the granularity of pruning, affects the efficiency of hardware accelerator design as well as the prediction accuracy. Coarse-grained pruning brings more regular sparsity patterns, making it more amenable for hardware acceleration, but more challenging to maintain the same accuracy. In this paper we quantitatively measure the trade-off between sparsity regularity and the prediction accuracy, providing insights in how to maintain the accuracy while having more structured sparsity pattern. Our experimental results show that coarse-grained pruning can achieve similar sparsity ratio as unstructured pruning given no loss of accuracy. Moreover, due to the index saving effect, coarse-grained pruning is able to obtain better compression ratio than fine-grained sparsity at the same accuracy threshold. Based on the recent sparse convolutional neural network accelerator (SCNN), our experiments further demonstrate that coarse-grained sparsity saves $\sim2\times$ of the memory references compared with fine-grained sparsity. Since memory reference is more than two orders of magnitude more expensive than arithmetic operations, the regularity of sparse structure leads to more efficient hardware design.
\end{abstract}

\section{Introduction}

Deep Neural Networks (DNNs) have many parameters, which leads to problems related to storage, computation and energy cost. State-of-art Convolutional Neural Network (CNN) models have hundreds of millions parameters and take tens of billions operations\cite{he2016deep,krizhevsky2012imagenet,simonyan2014very}. That makes DNN models difficult to deploy on embedded systems with limited resources.

To deal with this problem, various methods have been proposed to compress DNN models and reduce the amount of computation. Some methods are based on decomposition and factorization\cite{zhang2016accelerating,lebedev2014speeding}. These methods can preserve the regular dense computation structure of the original models, thus are able to to achieve both compression and acceleration on general-purpose processors. Pruning serves as another effective method to greatly reduce the number of parameters with no loss of accuracy\cite{han2015learning,guo2016dynamic}. 

Pruning based methods are better at preserving accuracy as well as achieving higher compression rates\cite{han2015learning}. However, such improvements come at the cost of the irregularity of the sparse computation pattern. On the other side, structured pruning, such as pruning entire filters will cause larger accuracy loss than pruning individual weights\cite{li2016pruning}. Those observations pose several questions: \emph{What is the trade-off between regularity and accuracy? Is it possible to find a sweet spot in the range of regularity? How does the sweet spot improve the efficiency of hardware implementation?}

\begin{figure}[t]
\centering
\vspace{-15pt}
\includegraphics[width=0.8\textwidth]{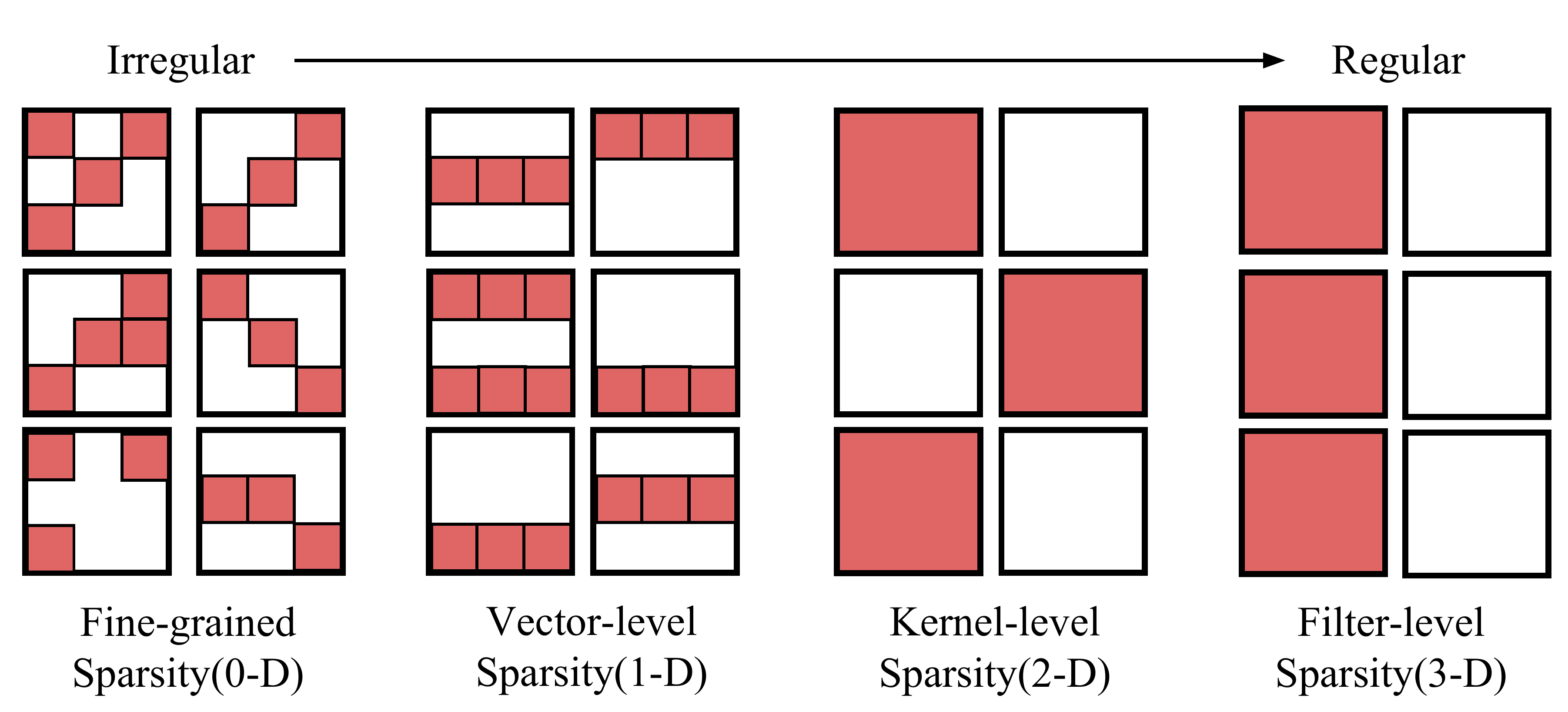}
\caption{Different sparse structure in a 4-dimensional weight tensor. Regular sparsity makes hardware acceleration easier.}
\label{fig:sparsity}
\vspace{-10pt}
\end{figure}

We attempt to answer those questions by looking into pruning with different granularity, as shown in Figure~\ref{fig:sparsity}. There are existing works trying to prune filters or channels instead of individual weights\cite{MolchanovPruningICLR17,wen2016learning,anwar2015structured}. However, they are individual points in the design space. Due to the various methods they used, we cannot directly evaluate the relationship between pruning granularity and final accuracy. 
We therefore apply the exact same method and experimental setting for an effective comparison. We also want to explore a consistent range of granularity, which includes intermediate grain size like 2D kernels and 1D sub-kernel vectors. Based on a thorough space exploration, we are able to analyze the storage saving and hardware efficiency at different granularity of sparsity.

In this work, we make the following contributions:
\begin{itemize}
\item We explore a complete range of pruning granularity and evaluate the trade-off between the model's regularity and accuracy. 
\item We demonstrate that coarse-grained pruning is able to reach similar or even better compression rates than the fine-grained one, even though it obtains less sparsity.
\item We show that coarse-grained sparsity is able to skip computations and reduce memory references in a structured manner, which leads to more efficient hardware accelerator implementation.
\end{itemize}

\section{Related Works}
\textbf{Methods of pruning}. Sparsity has been proven as an effective approach to save parameters of Deep Neural Network models\cite{han2015learning,guo2016dynamic}. A number of works have investigated how to select the important connections and effectively recover accuracy. Second-order derivative\cite{lecun1989optimal}, absolute value\cite{han2015learning}, loss-approximating Taylor expansion\cite{MolchanovPruningICLR17}, and output sensitivity\cite{engelbrecht2001new} are examples of importance metrics used for pruning. There are also methods trying to better integrate pruning and training, like iterative pruning\cite{han2015learning} and dynamic pruning\cite{guo2016dynamic}. 

\textbf{Granularity of sparsity}. Among all types of sparsity, fine-grained sparsity (vanilla sparsity) and filter-wise sparsity (very coarse-grained sparsity) are two extreme cases that has been studied\cite{han2015learning,li2016pruning}. Fine-grained sparsity is a type of sparsity in which individual weights are deleted and was first proposed in 1989 by LeCun et al.\cite{lecun1989optimal}. Fine-grained sparsity has been proven to work well on a wide range of popular neural network models of CNN and RNN\cite{han2015learning,guo2016dynamic,giles1994pruning,han2017ese}. There is also channel reduction and filter reduction, which  reduce the dimension of input/output features as well as layers. Channel reduction can be viewed as very coarse-grained sparsity that removes 3-dimensional sub-tensors in convolutional layers. Such coarse-grained sparsity is beneficial for acceleration due to regularity\cite{lebedev2016fast,wen2016learning}. 
%
%
%
%
However, it usually causes notable reduced accuracy compared with fine-grained sparsity, as indicated by Li et al.\cite{li2016pruning}.
%
%

There is a large range of granularity between vanilla sparsity and channel reduction. Some literature attempts to explore one or a few possibilities among all choices. Intra-kernel strided pruning is one case  investigated in the work of Anwar et al.\cite{anwar2015structured}.

\textbf{Accelerating sparse models}. For very coarse-grained sparsity like filter-sparsity and channel-sparsity, it is simple to achieve acceleration on general-purpose processors because it is equivalent to obtaining a smaller dense model\cite{wen2016learning}. For fine-grained sparsity, custom accelerators\cite{han2016eie,parashar2017scnn} have been proposed to exploit the reduction of computations.
%
%

\section{Granularity of Sparsity}
\subsection{Notations}
To simplify descriptions, we use the following notations for CNN. In a single convolutional layer, the weights compose a 4-dimensional tensor of shape $C\times K \times R \times S$. $C$ is the output dimension, i.e., the number of output feature maps. $K$ is the input dimension. $R$ and $S$ are the shape of convolution kernels ($R$=3, $S$=3 for a 3x3 kernel).

One layer's weights consist of multiple filters~(3-dimensional tensor of shape $K \times R \times S$), each one associated with an output feature map. The weights can also be viewed as multiple channels (3-dimensional tensor $C\times R \times S$), each one associated with an input feature map. Filters and channels are both composed of kernels (2-dimensional tensor $R \times S$), which are the key element in the 2-d convolution operation. Sub-kernel vectors (1-dimensional tensor of size $R$ or $S$) and scalar weights(0-dimensional tensor) are lower-level elements in a convolutional layer. Figure~\ref{fig:illu} illustrates these concepts.

\subsection{Range of Granularity}

It has been stated that fine-grained sparsity and channel reduction are two extreme cases of granularity. Among all possible grain sizes, there are a large variety of choices.

\begin{figure}[h!]
\centering
\noindent
\begin{minipage}{.43\linewidth}
\includegraphics[width=\textwidth]{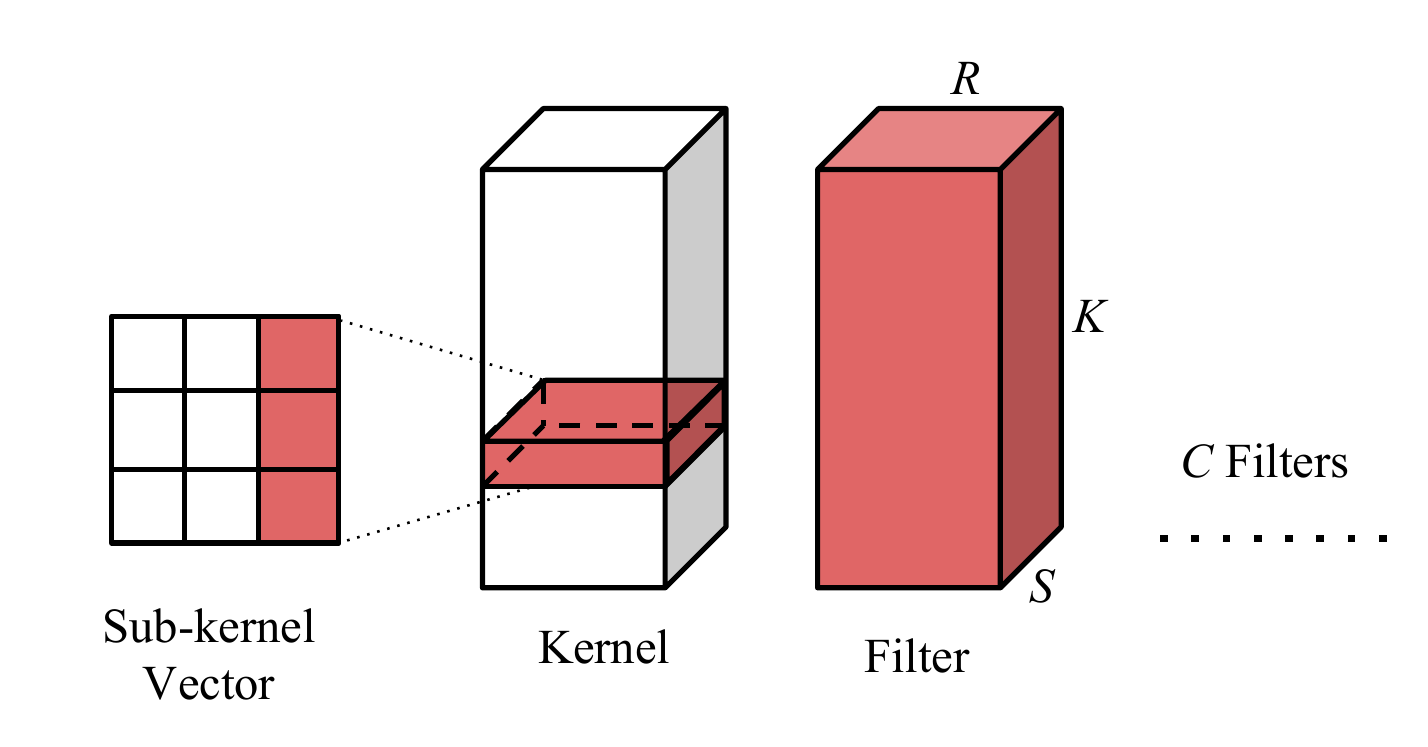}
\caption{Example of Sub-kernel Vector, Filter and Kernel.}
\label{fig:illu}
\end{minipage}%
\begin{minipage}{.49\linewidth}
\includegraphics[width=\textwidth]{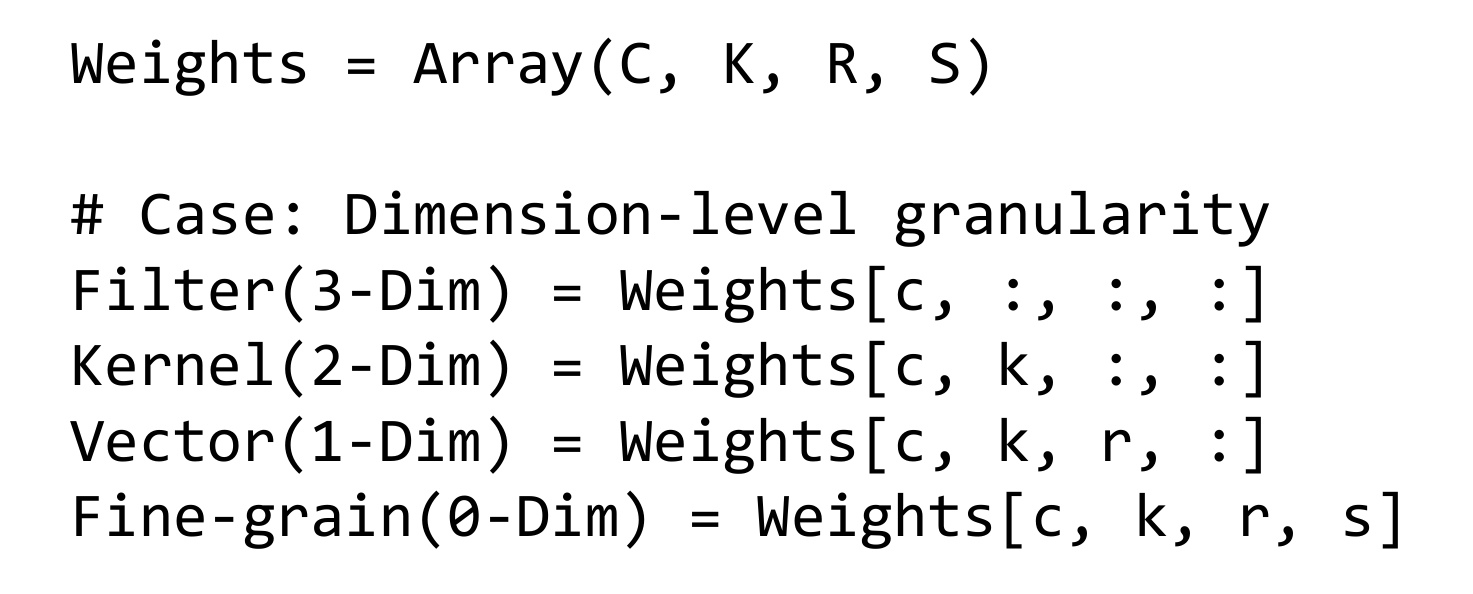}
\captionsetup{labelformat=empty}
\caption{Pseudo code: different granularity levels}
\label{pusedo_code}
\end{minipage}
\end{figure}

In this paper, we investigate granularity where the grain size increases with the number of dimensions.  To be specific, we study 4 cases where the atomic elements(grain) during pruning are scalar weights~(0-D), sub-kernel vectors~(1-D), kernels~(2-D) and filters~(3-D).
%
%
We explain these cases with numpy-like pseudo codes as below. Figure~\ref{fig:illu} also illustrates the different granularities of sparsity.
%
%
%
%

We select these four cases because they can be and have already been mapped into simple computation logics. 
\emph{Fine-grained sparsity} breaks the tensor into disjoint weights, therefore fine-grained multiplication-accumulation are required. Its implementations has been provided in EIE\cite{han2016eie} and SCNN\cite{parashar2017scnn}.  
%
%
\emph{Sub-kernel Vector sparsity} can be mapped into 1-D convolution. Though the sparse case has yet been studied, Eyeriss\cite{chen2016eyeriss} deals with the dense case and treats 1-D convolution as primitive operation.
\emph{Kernel sparsity} is related with 2-D convolution, which is primitive operations in a variety of algorithms and platforms, like Winograd convolution in cuDNN\footnote{\href{https://devblogs.nvidia.com/parallelforall/optimizing-recurrent-neural-networks-cudnn-5/}{Nvidia developer's blog}} and some FPGA implementations\cite{qiu2016going}. 
\emph{Channel sparsity} is equivalent to model reduction and, thus, is easy for acceleration in all types of platforms.

\subsection{Coarse-grained Pruning Method}

Coarse-grained pruning deletes multiple weights together instead of individual weights. Typically the grain can be filters, kernels, sub-kernels vectors or anything else. Because we are interested in the effects of the grain size rather than the pruning method, we adopt the simple magnitude-based pruning criterion in \cite{han2015learning}. For a grain $G_i$ that consists of multiple weights, the Salience $S_i$ is defined as the sum of absolute values, i.e. the L1 norm in Equation~\ref{equa:saliency}. Given the targeted sparsity, say we want 30\% of the weights to be zero, we sort the grains of weights according to the L1 norm defined in Equation~\ref{equa:saliency}. The grains with the smallest 70\% L1-norm are deleted.

\begin{equation}
\label{equa:saliency}
S_i = \sum_{w \in G_i} |w|
\end{equation}

We also adopt the iterative pruning method proposed by Han et al.\cite{han2015learning}. It is able to reach higher sparsity than direct pruning. 
The sparsity during each pruning stage is determined by sensitivity analysis, which requires individually pruning every layer and measure the accuracy loss on the training dataset.

\section{Sparsity-Accuracy Relation with Different Grain Sizes}
\begin{figure}[b!]
\centering
\includegraphics[width=0.75\textwidth]{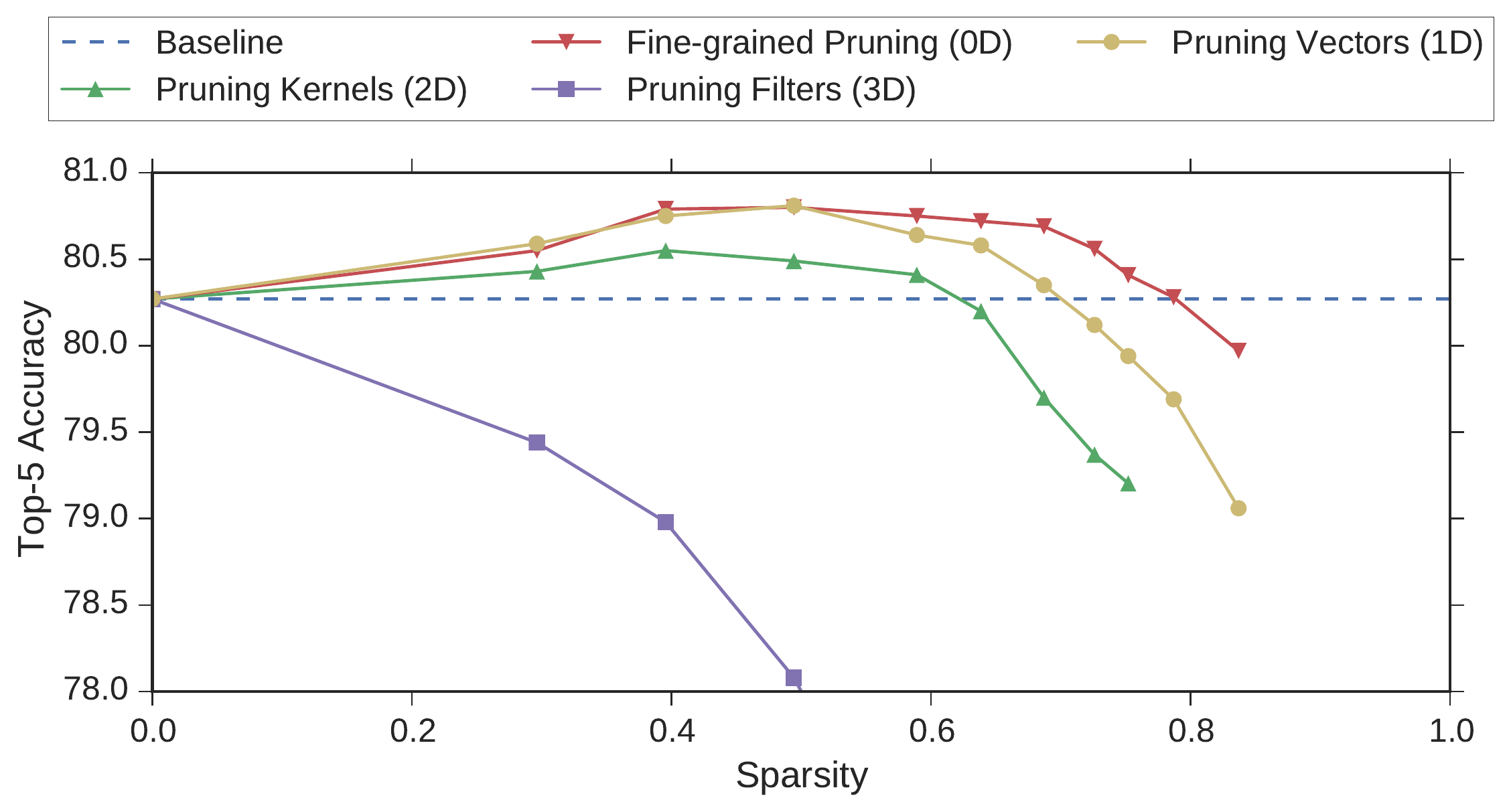}
\caption{Accuracy-Sparsity Curve of AlexNet obtained by iterative pruning. }
\label{accu-spar-alexnet}
\end{figure}

Our goal is to study how the granularity of pruning influences the accuracy. Specifically, we want to compare the accuracy of different pruning granularities at the same sparsity, and the result is shown in Figure~\ref{accu-spar-alexnet}. Sparsity serves as a regularizer because it lowers the model capacity by reducing the number of parameters. Coarse-grained sparsity not only reduces the number of parameters but also constrains the positions of parameters, which is an even stronger regularizer. That's why at low sparsity rate we  observed the accuracy improvement in Figure~\ref{accu-spar-alexnet}.
%
%
%
%
%
%

To ensure fair comparison, we enforce the same sparsity setting and training schedule for the same model. All experiments were performed on the ImageNet dataset\cite{deng2009imagenet} with Caffe\cite{jia2014caffe}.
%
%

For CNN models, we only count the overall sparsity of convolutional layers since convolutional layers take up most of the computations in a typical CNN model\cite{bagherinezhad2016lcnn}. However, we still prune the fully-connected layers together with convolutional layers to obtain consistent comparisons with previous works\cite{han2015learning,guo2016dynamic}. For fc layers, we only use fine-grained pruning because there's no such hierarchy of pruning granularity in the fc layer. 
%
%
%
%

We experimented on AlexNet for detailed accuracy-sparsity curves. We also experimented with modern CNNs, including VGG-16\cite{simonyan2014very}, GoogLeNet\cite{szegedy2015going}, ResNet-50\cite{he2016deep}, and DenseNet-121\cite{huang2016densely} and compared their accuracies at the same-sparsity point. Their results are reported and compared in Table~\ref{tab:comp_sparsity}. 

Figure~\ref{accu-spar-alexnet} shows the accuracy curve of density(1-sparsity) under various settings. In this figure there are four different types granularity of sparsity; in each case the atomic element for pruning is
\begin{itemize}
\item \textbf{Fine-grained(0-Dim)}: Individual weights.
\item \textbf{Vector(1-Dim) }: Sub-kernel vectors of size $S$.
\item \textbf{Kernel(2-Dim)}: Kernels of shape $R\times S$.
\item \textbf{Filter(3-Dim)}: Filters of shape $K\times R\times S$.
\end{itemize}

When the grain size of pruning is very large, say, filters, we observed huge accuracy loss during iterative pruning. AlexNet loses nearly 1\% validation accuracy at the very first pruning stage, which implies it is unsuitable for lossless model compression. For finer-grained pruning, the accuracy loss is much smaller; we even noticed small accuracy increases during the first several pruning stages. Note that the results for AlexNet are better than the original work by Han et al.\cite{han2015learning} due to a smoother pruning process. We give a detailed description in Section~\ref{sec:summary}.

\begin{table}[t]
\centering
\caption{Comparison of accuracies with the same density/sparsity.}
\label{tab:comp_sparsity}
\begin{tabular}{c|c|cc}
\hline
Model & Density & Granularity  & Top-5 \\
\hline
\multirow{3}{*}{AlexNet} & \multirow{3}{*}{24.8\%}& Kernel Pruning (2-D)   & 79.20\% \\
& &  Vector Pruning (1-D)  & 79.94\% \\
& & Fine-grained Pruning (0-D)  & \bf{80.41\%} \\
\hline
\multirow{3}{*}{VGG-16} & \multirow{3}{*}{23.5\%}& Kernel Pruning  (2-D)   & 89.70\% \\
& & Vector Pruning (1-D)  & 90.48\% \\
& & Fine-grained Pruning (0-D)  & \bf{90.56\%} \\
\hline
\multirow{3}{*}{GoogLeNet} & \multirow{3}{*}{38.4\%}& Kernel Pruning  (2-D)   & 88.83\% \\
& & Vector Pruning  (1-D)  & 89.11\% \\
& & Fine-grained Pruning (0-D)  & \bf{89.40\%} \\
\hline
\multirow{3}{*}{ResNet-50} & \multirow{3}{*}{40.0\%}& Kernel Pruning  (2-D)   & 92.07\% \\
& & Vector Pruning  (1-D)  & 92.26\% \\
& & Fine-grained Pruning (0-D)  & \bf{92.34\%} \\
\hline
\multirow{3}{*}{DenseNet-121} & \multirow{3}{*}{30.1\%}& Kernel Pruning  (2-D)   & 91.56\% \\
& &  Vector Pruning (1-D)  & 91.89\% \\
& & Fine-grained Pruning (0-D)  & \bf{92.21\%} \\
\hline
\end{tabular}
\end{table}

The results in Table~\ref{tab:comp_sparsity} and Figure~\ref{accu-spar-alexnet} support the assumption that coarse-grained sparsity causes greater accuracy loss than fine-grained sparsity.
Pruning with a large grain size like filters will greatly hurt accuracy. On the other hand, pruning with a smaller grain size leads to similar accuracy-sparsity curves with fine-grained pruning. Notice that in Figure~\ref{accu-spar-alexnet}, some curves appear to rise smoothly at first. That suggests coarse-grained pruning can still reach similar compression rates as fine-grained pruning, giving additional advantages described in the following section.

\section{Comparison of Storage}
\label{sec:storage}

\begin{figure}[h]
\vspace{-7pt}
\centering
\noindent\begin{minipage}{.45\linewidth}
\includegraphics[width=\textwidth]{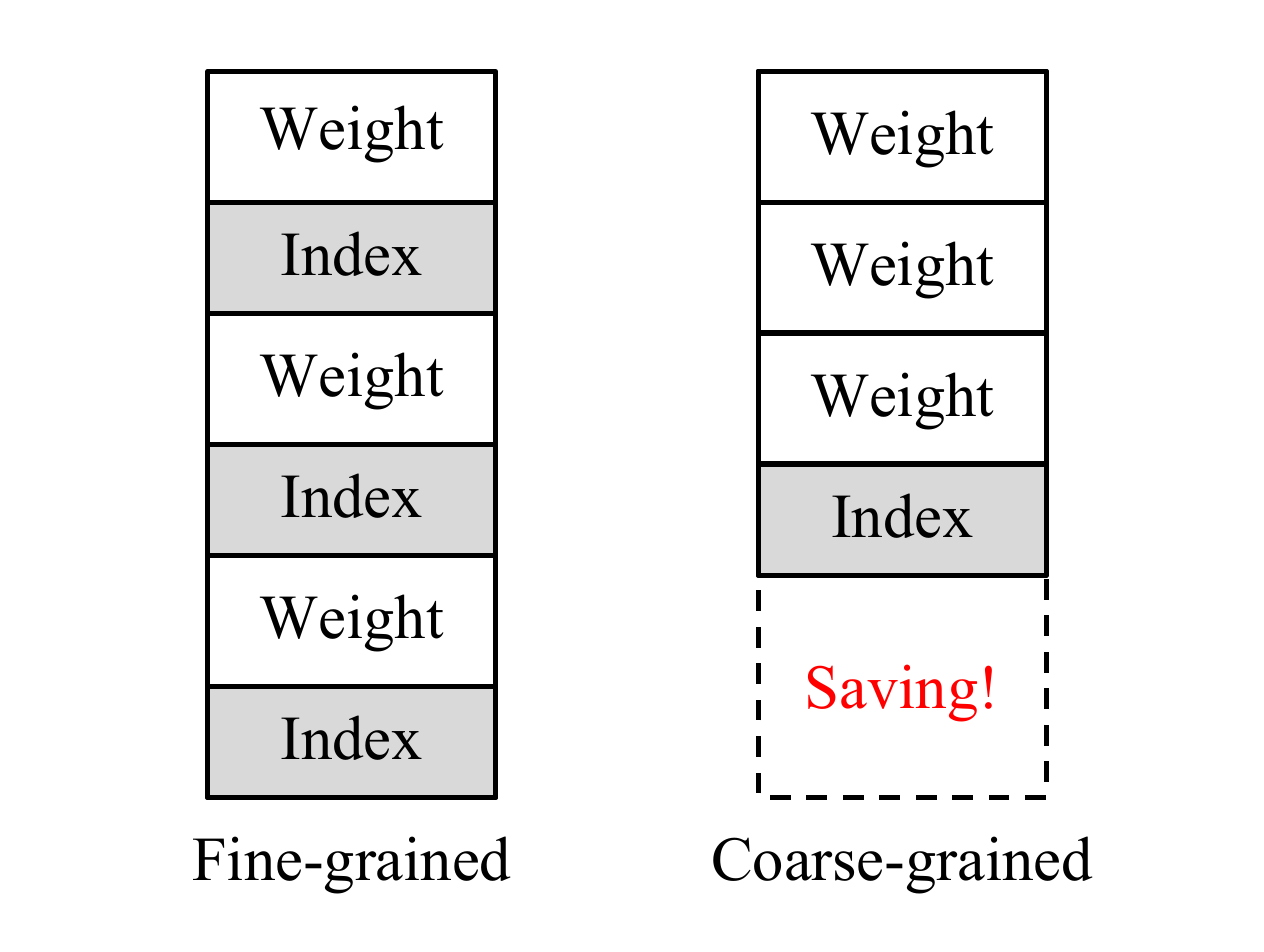}
\caption{Illustration of index saving.}
\label{fig:sav}
\end{minipage}%
\begin{minipage}{.43\linewidth}
\includegraphics[width=\textwidth]{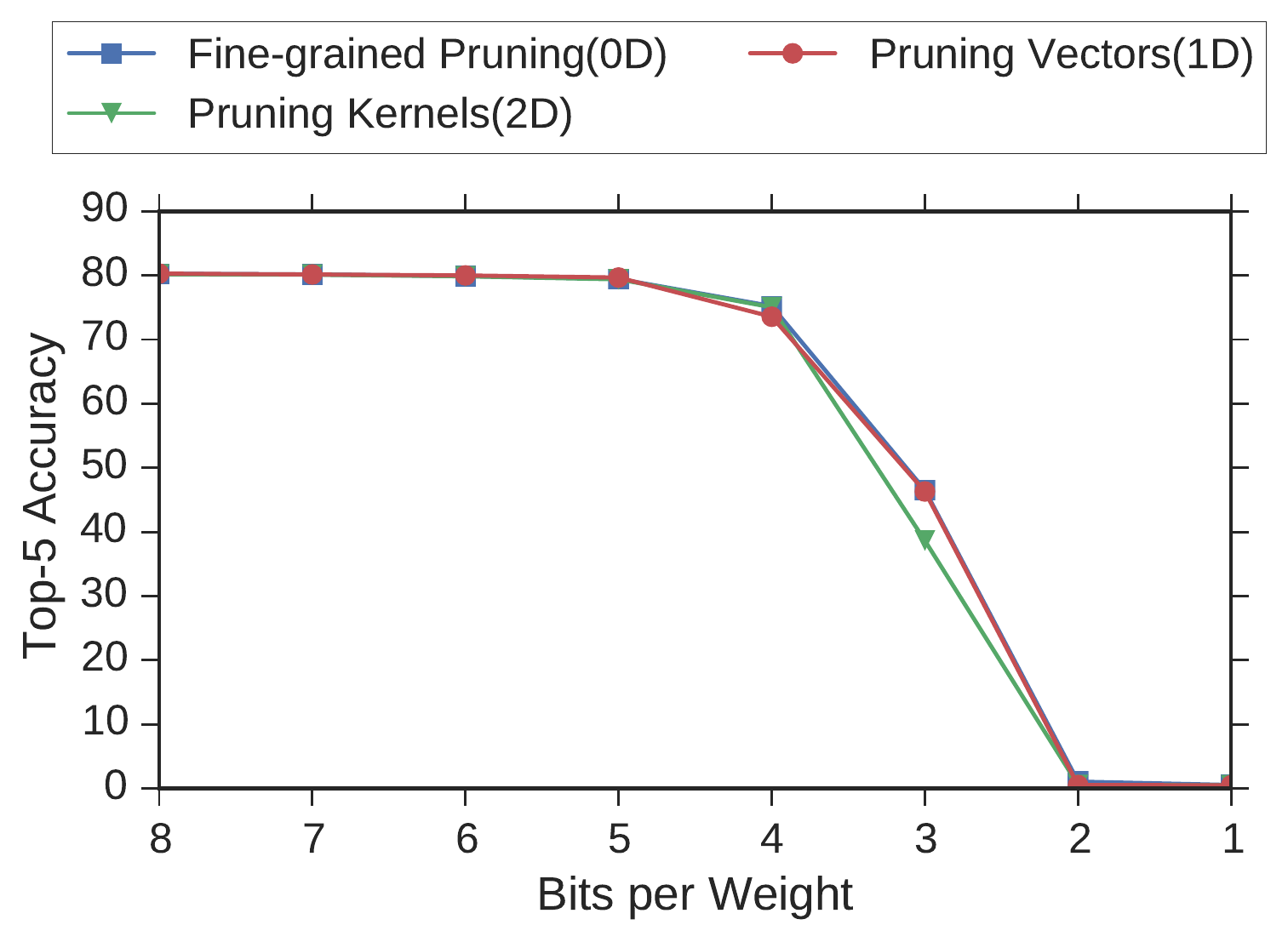}
\caption{Three curves are almost identical, indicating sparsity structure does not impact quantization.}
\label{fig:deep_comp}
\end{minipage}
\end{figure}

Model size is an important factor for real-world mobile applications. On the one hand, it constrains the application in memory-bounded devices. On the other hand, memory access is more than two orders of magnitude more energy expensive during the execution of deep neural network\cite{han2015learning}. Sparsity serves as an effective approach to compress neural network models. Sparse neural networks are usually stored in a similar format to Compressed Row Storage(CRS), where both values and indices are stored. Coarse-grained sparsity, due to its regularity, is able to save the number of indices as illustrated in Figure~\ref{fig:sav}. Therefore the coarse-grained sparse models take up less storage than fine-grained models at the same sparsity.
%
%

\begin{figure}[t]
\centering
\includegraphics[width=0.76\textwidth]{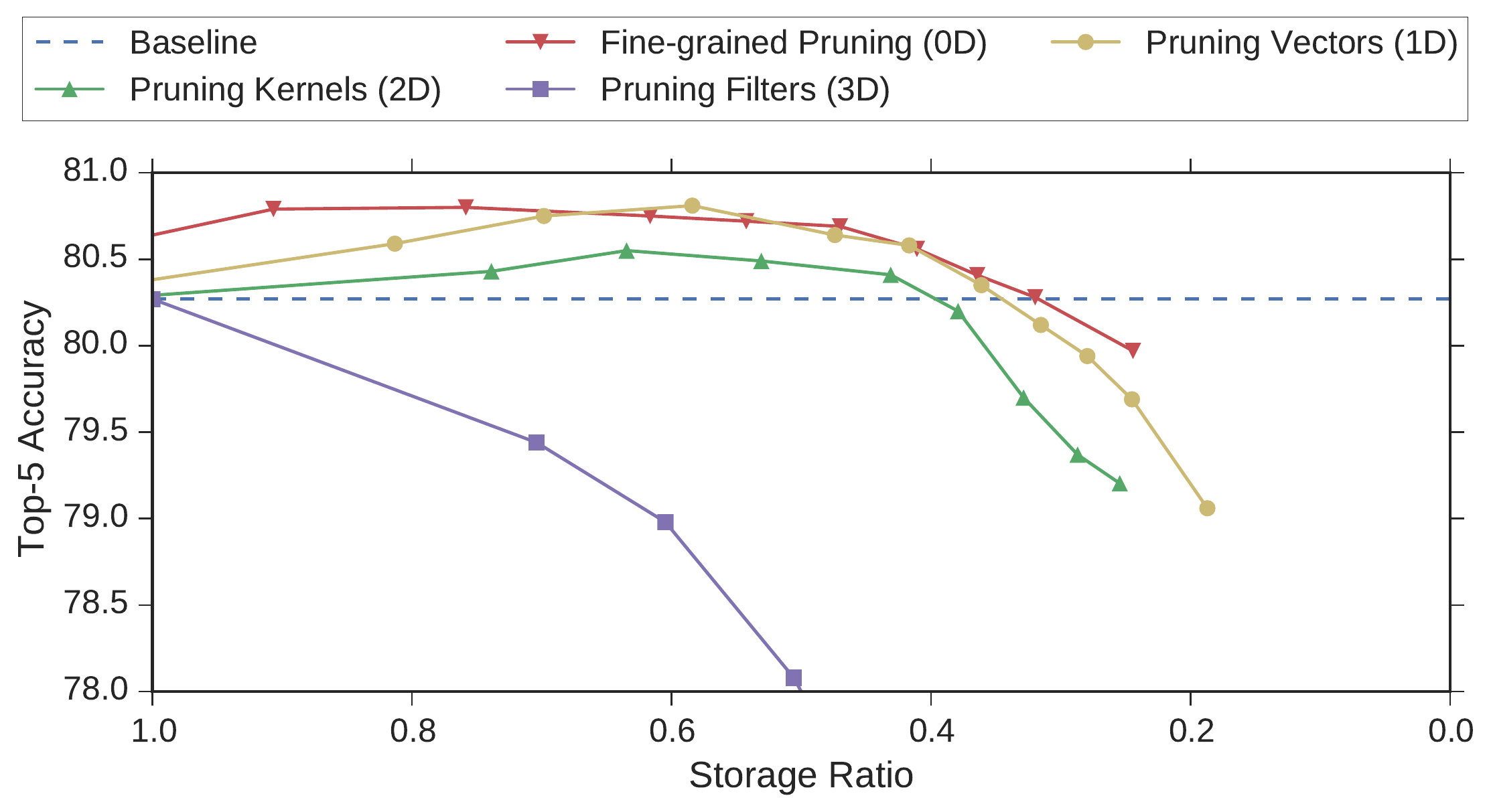}
\caption{Accuracy-Storage Curve of AlexNet with different grain sizes. Notice that vector pruning only causes 1.5\% more storage and kernel pruning causes 6.7\% more storage.}
\label{accu-storage}
\end{figure}

We want to investigate how the prediction accuracy changes with different grain sizes of pruning at the same level of storage(instead of sparsity). We do not use full-precision 32-bit weights, but 8-bit weights instead, as 8-bit weights, either true 8-bit integer formats or 8-bit indices indexing to a table of shared fp32 weights, have been proven to be sufficient in a lot of literature\cite{Jouppi2017,han2016eie,vanhoucke2011improving}.
%
%
%
%
We use 4-bit indices to store the distances between adjacent non-zeros, following the method in Deep Compression \cite{han2015deep}. Moreover, as indicated in Deep Compression, the quantization method works independently with sparsity. To check if it still works with coarse-grained sparsity, we plot the accuracy-bits curves of different types of pruned models in Figure~\ref{fig:deep_comp}. The results show that sparsity structure has negligible influence over quantization.
%
%
%
%

Figure~\ref{accu-storage} shows the accuracy-storage relationship of AlexNet. We find that the first three curves(Fine-grained, Vector and Kernel) are closer than those in Figure~\ref{accu-spar-alexnet}. This figure shows the effect of index saving for coarse-grained pruning.
%
%

To better compare the compression ratio with modern deep neural nets, we list the results of AlexNet, VGG-16 and GoogLeNet, ResNet-50 and DenseNet-121 in Table~\ref{tab:comp_compress}. Here the storage ratio is defined as the model size of pruned 8-bit models(with 4-bit indices) to that of dense 8-bit models. Note that it is almost impossible to prune a model that exactly matches the baseline accuracy, so we use linear interpolation to obtain the estimated density and storage ratio at a given point of accuracy.

\begin{table*}[h!]
\centering
\caption{Comparison of storage savings at the baseline accuracy. Storage ratio is compared with the 8-bit dense model. 
}
\label{tab:comp_compress}
\begin{tabular}{c|c|p{3.8cm}p{2cm}c}
\hline
\multirow{2}{*}{Model} & Top-5   & \multirow{2}{*}{Granularity}  & \multirow{2}{*}{Density}    & \multirow{2}{*}{Storage Ratio} \\
& Accuracy &  &     &    \\
\hline
\multirow{3}{*}{AlexNet} & \multirow{3}{*}{80.3\%}& Kernel Pruning (2-D)  & 37.8\% & 39.7\% \\
& & Vector Pruning (1-D) & 29.9\% & 34.5\% \\
& & Fine-grained Pruning (0-D) &  22.1\% & \bf{33.0\%} \\
\hline
\multirow{3}{*}{VGG-16} & \multirow{3}{*}{90.6\%}& Kernel Pruning (2-D)& 44.4\% & 46.9\% \\
& & Vector Pruning (1-D) & 30.7\% & \bf{35.8\%} \\
& & Fine-grained Pruning (0-D) &  27.0\% & 40.6\% \\
\hline
\multirow{3}{*}{GoogLeNet} & \multirow{3}{*}{89.0\%}& Kernel Pruning (2-D) & 43.7\% & 51.6\% \\
& & Vector Pruning (1-D) & 36.9\% & \bf{47.4\%} \\
& & Fine-grained Pruning (0-D) &  32.3\% & 48.5\% \\
\hline
\multirow{3}{*}{ResNet-50} & \multirow{3}{*}{92.3\%}& Kernel Pruning (2-D) & 61.3\% & 77.0\% \\
& & Vector Pruning (1-D) & 40.0\% & \bf{52.7\%} \\
& & Fine-grained Pruning (0-D) &  37.1\% & 55.7\% \\
\hline
\multirow{3}{*}{DenseNet-121} & \multirow{3}{*}{91.9\%}& Kernel Pruning (2-D) & 35.5\% & 48.9\% \\
& & Vector Pruning (1-D) & 31.1\% & 43.8\% \\
& & Fine-grained Pruning (0-D) &  26.6\% & \bf{39.8\%} \\
\hline
\end{tabular}
\end{table*}

For a sparse network, the larger the grain size is, the less storage needed. This is due to index sharing among the weights of the kernel as shown in Figure~\ref{fig:sav}. However, AlexNet and VGG-16 in particular have much closer density/storage results for kernel pruning than GoogLeNet, ResNet, and DenseNet. This is caused by the small size of the convolutional kernels being pruned: these networks have many layers of 1x1 convolutions, which do not benefit from sharing index values. AlexNet and VGG-16, on the other hand, have a multitude of larger convolutions.

\section{Regular Sparsity Helps Hardware Implementation}
It has been mentioned in the previous sections that filter pruning is able to obtain acceleration on general-purpose processors like CPUs or GPUs. For intermediate grain sizes like kernels or sub-kernel vectors, though it is still difficult for acceleration on general-purpose processors, there are several advantages over fine-grained sparsity on customized hardware. Those advantages enable simpler circuit design and higher energy efficiency on customized hardware. We qualitatively and quantitatively analyze the advantages as follows:

\textbf{Qualitative analysis}. In convolutional layers, 2-D convolution is usually the primitive operation. 
Kernel pruning (2-D pruning) can easily lead to computation reduction, because the 2-D convolutions of deleted kernels can be saved. Recent custom hardware design for CNN also use 1-D convolution as the primitive operation\cite{chen2016eyeriss}. In this case, sub-kernel vector pruning is beneficial. Compared with fine-grained sparsity, coarse-grained sparsity is able to preserve the low-level computation logic, therefore simplifying the hardware design.

\begin{figure}[h]
\centering
\includegraphics[width=0.65\textwidth]{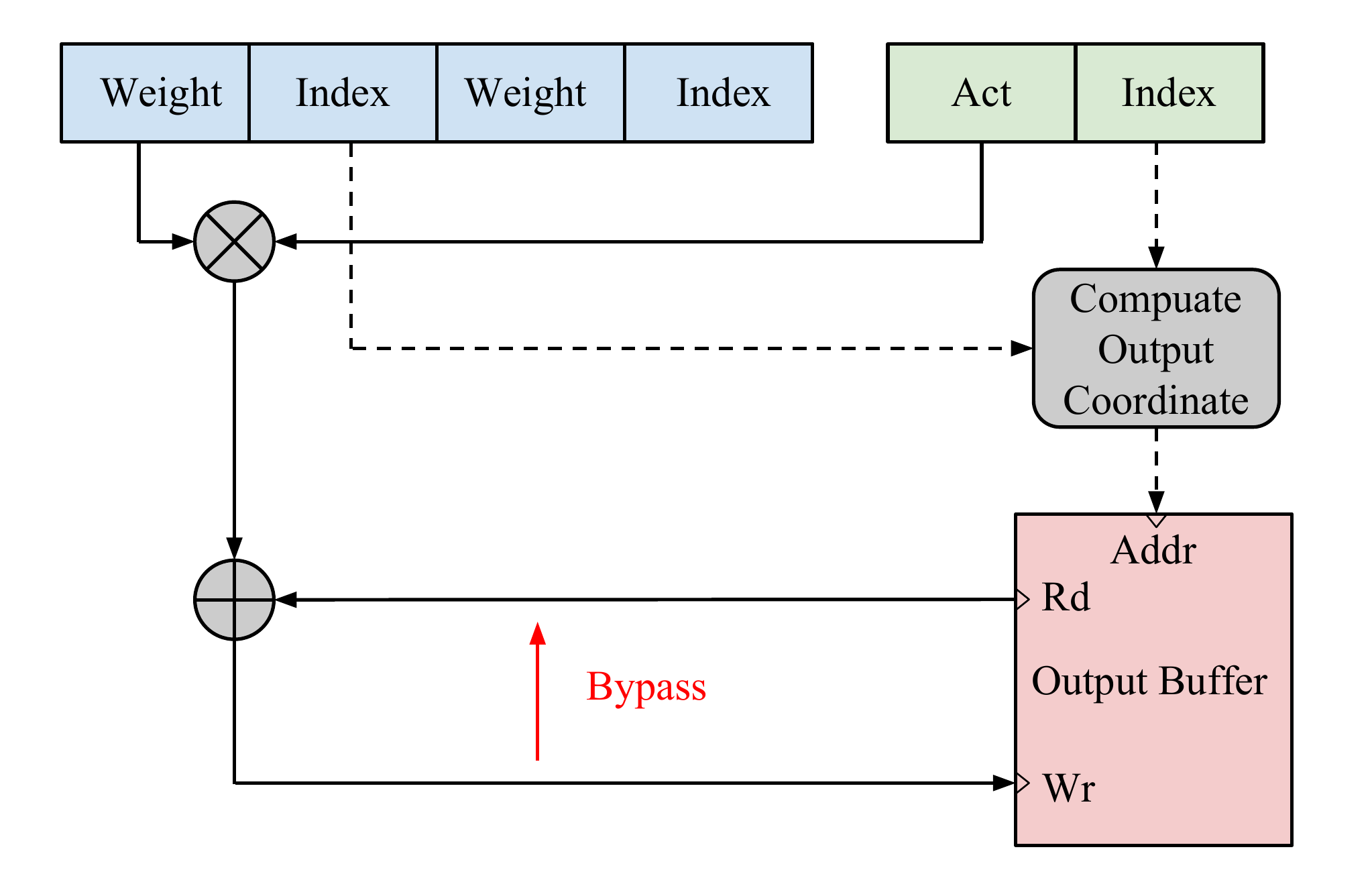}
\caption{A simplified dataflow of SCNN architecture. Weights and activations are both stored in sparse format. Bypass is possible when the same output address is referenced again.}
\label{fig:scnn}
\end{figure}

\textbf{Quantitative analysis}. Memory reference is a major factor of energy consumption\cite{han2015learning}. Recent work on custom hardware exploits both the sparsity of weights and activations of CNNs\cite{parashar2017scnn}. In their implementation, the weights and input activations are both stored in sparse format while output activations are stored in dense format. The indices of weights and activations are used for calculating the output address, to which the product of weight and activation will perform a scatter-add operation. This process is illustrated in Figure\ref{fig:scnn}. After one layer is finished, the output activations will then be compressed into the sparse format for next layer.

If the same output address is referenced again, a data shortcut can be used to avoid the expensive read/write. For example, two adjacent weights and two adjacent activations will reference 3 addresses instead of 4. Due to the locality, coarse-grained sparse weights have a larger probability of output address collision. We simulated with VGG-16 on ImageNet's validation set to compare the number of memory references and listed the results in Table~\ref{tab:memref}. With the same density, coarse-grained sparsity saves $30\%-35\%$ of the total output memory references.
%
%

\begin{table}[h!]
\centering
\caption{Output memory references for VGG-16 (convolutional layers only).}
\label{tab:memref}
\begin{tabular}{c|cc|c}
\hline
\multirow{2}{*}{Density} & Fine-grained & Vector Pruning  & \multirow{2}{*}{Relative \# of memory references} \\
& (0-D) & (1-D) & \\
\hline
40.1\% & 1.77B & 1.23B & \bf{69.5\%} \\
33.1\% & 1.53B & 1.03B & \bf{67.2\%} \\
27.5\% & 1.33B & 0.87B & \bf{65.3\%} \\

\hline
\end{tabular}
\end{table}

\section{Summary}
\label{sec:summary}
In this section we compare our results with previous works on pruning\cite{han2015learning,guo2016dynamic}. We select AlexNet, as its layer-wise sparsity is published in previous papers. By using a smoother pruning process, we find the results of Han et al.\cite{han2015learning} can be further improved without any algorithmic change. 

Table~\ref{tab:alexnet} gives an overall comparison of key statistics for AlexNet.  Apart from the number of parameters, there are some other factors affecting the efficiency of DNN models. Here FLOPs is the total number of floating-point operations. Storage is measured with of 8-bit weights and 4-bit indices as indicated in Section~\ref{sec:storage}. Due to the fact that the storage of convolutional layers is much smaller but reused much more frequently than fully-connected layers, we add an additional row for storage of convolutional layers. The number of memory referenced is calculated by simulating the process of Figure~\ref{fig:scnn}. Here the baseline number of memory references is obtained from dense model with sparse activations.
%
%

The results show that the our fine-grained pruned model already has advantages over the previous state-of-art work in terms of FLOPs, storage of convolutional layers and number of memory references. Moreover, compared with our fine-grained baseline, the vector pruning method can further reduce the storage of convolutional layers by 23\% and the number of memory references by 43\%.
%
%

\begin{table}[h!]
\centering
\caption{Comparison of pruned AlexNet with previous works which used fine-grained pruning. }
\label{tab:alexnet}
\begin{tabular}{c|c|ccccc}
\hline
\multirow{3}{*}{Layer} & \multirow{3}{*}{Param.} & NIPS'15  & NIPS'16   & Fine-grained & Vector & Kernel \\
& & \cite{han2015learning} & \cite{guo2016dynamic} & Pruning &  Pruning  &  Pruning  \\
& & & & (ours)& (ours) & (ours)\\
\hline
conv1 & 35K  & 84\% & \bf{54}\% & 83\% & 83\% & 83\%\\
conv2 & 307K & 38\% & 41\% & \bf{26}\% & \bf{26}\% & \bf{26}\%\\
conv3 & 885K & 35\% & 28\% & \bf{23}\% & \bf{23}\% & \bf{23}\%\\
conv4 & 664K & 37\% & 32\% & \bf{23}\% & \bf{23}\% & \bf{23}\%\\
conv5 & 443K & 37\% & 33\% & \bf{23}\% & \bf{23}\% & \bf{23}\%\\
fc6   & 38M  & 9\% & \bf{3.7}\% & 7\% & 7\% & 7\%\\
fc7   & 17M  & 9\% & \bf{6.6}\% & 7\% & 7\% & 7\%\\
fc8   & 4M  & 25\% & \bf{4.6}\% & 18\% & 18\% & 18\%\\
Total & 61M & 11\% & \bf{5.7}\% & 8.4\% & 8.4\% & 8.4\% \\
\hline
FLOPs  & 1.5B & 30\% & 25.4\% & \bf{24.1}\% & \bf{24.1\%} & \bf{24.1\%} \\
Storage(conv) & 2.3MB & 55.6\% & 48.3\% & 36.4\% & 28.0\% & \bf{25.5\%} \\
Storage(total) & 61MB & 16.7\% & \bf{8.5\%} & 12.6\% & 12.3\% & 12.2\% \\
\#Mem Reference & 99M & 74.4\% & 71.7\% & 60.5\% & \bf{34.6\%} & 35.2\%\\
\hline
\multicolumn{2}{c|}{Top-5 Accuracy} & 80.23\% & 80.01\% & \bf{80.41}\% & 79.94\% & 79.20\%  \\
\hline
\end{tabular}
\end{table}

\section{Conclusion}
We thoroughly explored the granularity of sparsity with experiments on detailed accuracy-density relationship. Due to the advantage of index saving, coarse-grained pruning is able to achieve a higher model compression ratio, which is desirable for mobile implementation. We also analyzed the hardware implementation advantages and show that coarse-grained sparsity saves $\sim 2\times$ output memory access compared with fine-grained sparsity, and $\sim 3\times$ compared with dense implementation. Given the advantages of simplicity and efficiency from a hardware perspective, coarse-grained sparsity enables more efficient hardware architecture design of deep neural networks.

{\small
\bibliographystyle{unsrt}
\bibliography{nips_2017}
}

\end{document}